\documentclass[letterpaper, 10 pt, conference]{ieeeconf}  % Comment this line out if you need a4paper

\IEEEoverridecommandlockouts                              % This command is only needed if 
\usepackage[english]{babel}       
\usepackage{pgf}

\usepackage{amsmath}                    % AMS Math Packet
\usepackage{amsfonts}                   % AMS Math Packet (Fonts)
\usepackage{amssymb}                    % Additional mathematical symbols
\usepackage{rotating}
\usepackage{booktabs}   
\usepackage{hyperref}% Nicer tables
\usepackage{cite}
\usepackage{acronym}                    % Acronyms
\usepackage{tikz} 
\usetikzlibrary{arrows,decorations.pathmorphing,backgrounds,fit,positioning,shapes.symbols,chains}
\usetikzlibrary{calc}
\title{\LARGE \bf
Model Mediated Teleoperation with a Hand-Arm Exoskeleton in Long Time Delays Using Reinforcement Learning
}

\author{Hadi Beik-Mohammadi$^{1,2,*}$, Matthias Kerzel$^{2}$, Benedikt Pleintinger$^{1}$, Thomas Hulin$^{1}$,\\ Philipp Reisich$^{1}$, Annika Schmidt$^{1,3}$, Aaron Pereira$^{1}$, Stefan Wermter$^{2}$ and Neal Y. Lii$^{1}$ % <-this % stops a space
\thanks{$^{1}$ Institute of Robotics and Mechatronics, German Aerospace Center, Munich, Germany
        {\tt\small (Hadi.Beikmohammadi, Thomas.Hulin, Annika.Schmidt, Benedikt.Pleintinger, Philipp.Reisich, Aaron.Pereira, Neal.Lii)@DLR.de}}%
\thanks{$^{2}$Knowledge Technology Institute, Dept of Informatics, Hamburg University, Hamburg, Germany
        {\tt\small (Kerzel, Wermter) @ informatik.uni-hamnburg.de}}%
\thanks{$^{3}$  Technical University of Munich, Munich, Germany
        {\tt\small (An.Schmidt)@TUM.de}}%
}
\begin{document}
\maketitle
\thispagestyle{empty}
\pagestyle{empty}

%%%%%%%%%%%%%%%%%%%%%%%%%%%%%%%%%%%%%%%%%%%%%%%%%%%%%%%%%%%%%%%%%%%%%%%%%%%%%%%%
\begin{abstract}
Telerobotic systems must adapt to new environmental conditions and deal with high uncertainty caused by long-time delays. As one of the best alternatives to human-level intelligence, Reinforcement Learning (RL) may offer a solution to cope with these issues.
This paper proposes to integrate RL with the Model Mediated Teleoperation (MMT) concept. The teleoperator interacts with a simulated virtual environment, which provides instant feedback. Whereas feedback from the real environment is delayed, feedback from the model is instantaneous, leading to high transparency.
The MMT is realized in combination with an intelligent system with two layers. The first layer utilizes Dynamic Movement Primitives (DMP) which accounts for certain changes in the avatar environment. And, the second layer addresses the problems caused by uncertainty in the model using RL methods. 
Augmented reality was also provided to fuse the avatar device and virtual environment models for the teleoperator.
Implemented on DLR's Exodex Adam hand-arm haptic exoskeleton, the results show RL methods are able to find different solutions when changes are applied to the object position after the demonstration. The results also show DMPs to be effective at adapting to new conditions where there is no uncertainty involved.

\end{abstract}

%%%%%%%%%%%%%%%%%%%%%%%%%%%%%%%%%%%%%%%%%%%%%%%%%%%%%%%%%%%%%%%%%%%%%%%%%%%%%%%%
\section{INTRODUCTION}
\label{section:intro}
%Final_Cut% %In a teleoperation scenario when there is a long distances between the teleoperator and commanded robot system, data transmission
%between the two sides 
%can take on significant time delay. This time delay introduces inconsistency or mismatch between input and avatar systems that falsifies the transmitted data. For example, the teleoperator may receive the haptic feedback well after the avatar robot has already collided with an object in the remote environment, causing damage the target robot system, as well as the object and environment it is interacting with. The visual information provided to the operator may also have significant discrepancies from the real-time action. In a grasping scenario, the object may change pose or position in the span of the communication delay, which can cause the teleoperated grasp to be compromised or fail. Therefore, relying exclusively on the state of the remote environment is undesirable.
% NL comment: Not sure we should call it a framework
Teleoperation provides the possibility for an operator to interact with a remote environment using an intermediate device. The intermediate device consists of two interconnected parts so-called input and avatar. Using the input device, the teleoperator remotely controls/commands the avatar through a communication channel. Although the avatar is assumed to be passive, due to the bilateral control scheme, it affects the input device using position or force feedback. The bilateral control scheme provides feedback to the operator, which augments the remote environment to be perceived by the operator's senses, such as haptic, visual, and auditory.
In a long-distance teleoperation scenario, data transmission can take a significant time delay to reach the teleoperator. This delay introduces inconsistency or mismatch between input and avatar system that falsifies the provided feedback. For example, the teleoperator may receive the haptic feedback well after the avatar robot has already collided with an object in the remote environment, causing damage to the target robot system, as well as the remote environment. Furthermore, the visual information provided to the operator may also have significant discrepancies. For example, the object may move in the span of the communication delay, which can cause the task to be compromised. Therefore, relying solely on the data collected from the remote environment is undesirable.
\begin{figure}[t]
% \vspace{2mm}
    \centering
    \includegraphics[width=\linewidth]{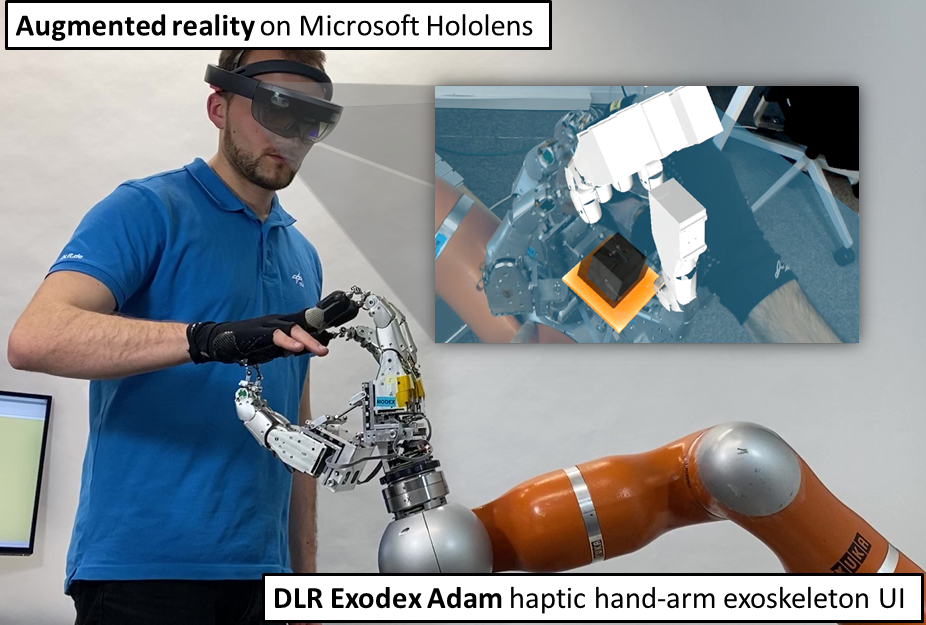}
    %\caption{The figure shows the teleoperator uses the Exodex Adam haptic interface to teach the avatar robot to build a tower using three cubes. The teleoperator observes a replica of remote environment and the avatar robot using the augmented reality.}
    \caption{A teleoperator using the proposed approach with the Exodex Adam hand-arm haptic interface and a Microsoft Hololens teaches the avatar robot to build a tower using three cubes. The teleoperator observes a replica of the remote environment and the avatar robot using the augmented reality.}
    \label{fig:first_image}
    \vspace{-6mm}
\end{figure}
\noindent

\noindent
Grasping or manipulation of objects by teleoperation can be realized in different ways. The robot configurations can be pre-recorded and used as a look-up table. The corresponding configurations can be selected based on the closest object position to the current position \cite{Hadi_ICRA}. To reduce the inconsistency between configurations, end-to-end deep learning with pre-trained neural networks \cite{Hadi_IROS} can be used instead of hard-coded look-up tables. Furthermore, deep reinforcement learning can be used as an online solution where the agent learns to reach for grasp by interacting with the environment \cite{Hadi_ICANN, Matthias_IJCNN, Matthias_grasp}. The idea of learning from demonstration (LfD) can increase the stability and performance of the grasping and in-hand manipulation \cite{freekpi2, mastersthesis1}.

\noindent
In practice, a teleoperation scenario may take place on different levels of abstraction, automation, and shared autonomy. Using a task-driven approach with gesture recognition, it has been demonstrated that in-hand manipulation can be effectively carried out for all six possible Degrees of Freedom (DOF), on free and partially constrained objects \cite{taskman}. German Aerospace Center (DLR) and European Space Agency's (ESA) METERON SUPVIS Justin \cite{justin2} was a teleoperation mission with high-level abstraction and complete task autonomy where an astronaut in orbit commanded a robot on the ground to execute a task. In this experiment, the task performance relied mainly on the avatar's built-in intelligence and the intuitiveness of the task representation for the teleoperator.
% The task also can be a combination of primitive tasks such as pulling, pushing, grasping, in-hand-manipulation and placing an object. 
%\color[rgb]{1,0,0}The astronauts were briefed about how to demonstrate the task in a high-level graphical user interface but none of them were informed about how the tasks were deployed on the low-level motion planners. In this case, the task performance relies mainly on avatar's intelligence while low-level manipulation tasks rely on the operator's skills, prior knowledge, and sufficient sensory feedback.\textbf{Can shrink this to a line. You need space. }\color[rgb]{0,0,0}

\noindent
Teleoperators continue to demand immersive user experience with haptic feedback of force reflection to feel the object and the environment. Various work has been carried out to study force reflection in-hand telepresence, including an exoskeleton system that uses neural networks~\cite{fischer1998learning}. 

\noindent
Furthermore, previous work \cite{dlr80501} has evaluated and shown the effectiveness of haptic feedback for increasing the in-hand teleoperation performance, as compared to other feedback conditions. By considering the viscosity of the environment, a telepresence system can be extended to interact in a mixed media environment of fluid, gas, and solids up to the fingertips to realize an even more immersive user experience \cite{mastersthesis_annika}. 

\noindent
%Final_Cut%
%Teleoperation at long distances involve transmitting data with time delay. The duration of the time delay depends on distance, communication bandwidth, network buffering, processing on relay stations or other factors. Therefore, the feedback may reach the operator with long time delays. 

\noindent
The bilateral teleoperation architecture usually contains 2-channel \cite{2_channel} or 4-channel \cite{4_channel} communication, which ensures the consistency between two models and the safety of the avatar.  
For example, if during a task an object in the remote environment hinders the avatar movement, instant force feedback can inform the operator to prevent colliding with an object or stop penetrating a hard surface. Instant force feedback requires high bandwidth communication channels with no time delay.

%The latency varies in different applications which can last a few milliseconds up to hundreds of seconds. 
\noindent
As mentioned, time delay can cause teleoperation to become unstable, particularly for high-DOF, long-delay systems. Kontur-2 \cite{dlr112967} and METERON Haptics-2 experiments\cite{Haptics2} have tackled this with 20 msec to 800 msec+ of time delays. In robot-assisted remote telesurgery, the operations are handled with delay as long as 100 msec+ \cite{Telesurgery}.

\noindent
Approaches such as passivity-based control \cite{passivity1, passivity2, passivity4} and predictive display \cite{predictivediplay1, predictivedisplay2} can reduce the inconsistency, but the problems appear when the time delay increases or becomes variable. A unique way of approaching the time delay problem is introduced by Model Mediated Teleoperation (MMT) approach \cite{modelmediated1, Xu2016ModelMediatedTT}. The MMT approach uses a virtual replica of the remote environment on the input-side to provide instant feedback for the teleoperator and has been used in several areas, for example, surgical teleoperation \cite{SurgicalMMT} and space robotics \cite{SpaceMMT_DLR}.

\noindent
MMT requires an accurate model of the remote environment and avatar device, but modeling a non-linear time-variant system like a remote environment is problematic. 
 The MMT approaches attempt to recreate the remote environment model using different methods such as neural networks \cite{iet:/content/books/ce/pbce037e}.
To tackle problems caused by the time delay, proving a form of timely visual and haptic feedback combined with adaptation to uncertainties is required.
The avatar-side intelligence can be realized to compensate for errors and mistakes in the task demonstration; therefore, discarding the need for an accurate model of the remote environment on the input-side. Hence, providing instant haptic and visual feedback does not require the precise future state of the environment, and a simple approximation is sufficient. 

\noindent
Our approach adopts the architecture from the MMT scheme and combines it with Dynamic Movement Primitives (DMP) \cite{MainDMP} and model-free Reinforcement Learning (RL) \cite{MainPI2, freekpi2, PowerAlgorithm}. The avatar system uses DMPs to account for changes in the model by transforming the trajectory into non-linear dynamical systems. The reconstructed trajectory can be generated almost instantly. To improve the adaptation, the system should be able to adapt to uncertainties in the model that cannot be considered before trajectory execution. The policy search RL methods are integrated to explore and search for a viable solution in a limited time. Three RL methods are evaluated, Policy learning by Weighting Exploration with the Returns (PoWER) \cite{PowerAlgorithm} based on expectation maximization, Episodic Natural Actor Critic (eNAC) \cite{enac} based on natural gradient, and Policy Improvement using Path Integrals ($\textrm{PI}^2$) \cite{MainPI2} based on stochastic optimal control. 

\noindent
%As already mentioned, 
Various forms of feedback can be used to immerse the teleoperator in the remote environment. As Fig. \ref{fig:first_image} shows, our system offers haptic feedback via the haptic Exodex Adam interface \cite{p:Exodex-Prime,p:Exodex-Adam}, in which all forces are calculated in the virtual environment. The teleoperator controls an anthropomorphic robot arm via the haptic interface, which is supposed to reproduce movements at the same time. Furthermore, a simulated robot farm using multiple instances of the avatar system and the remote environment is developed to facilitate the learning process. 
\noindent
The rest of the paper is organized as follows: Sec. \ref{section:approach} presents the proposed RL enhanced MMT approach and architecture of the teleoperation system. Sec. \ref{section:implementation} presents the design and the implantation of the necessary modules to provide a practical framework. In Sec. \ref{section:results}, the results are presented. In Sec. \ref{section:discussion} a general discussion about the approach is provided. Finally, Sec. \ref{section:conclusion} concludes the paper and lays out our future work.

%\section{BACKGROUND}
%
%\color[rgb]{1,0,0}\textbf{talk about some non-space scenarios, too. Your paper is not about space teleoperation, yet all your references are space stuff. You can talk about telesurgery, catastrophe rescue scenarios, etc. It is okay to swap out or shink the space stuff}\color[rgb]{0,0,0}

% \begin{figure}%[ht]
%     \centering
%     \includegraphics[width=\linewidth]{images/BasicArchitectureV2.pdf}
%     \caption[The figure shows the general structure of a telemanipulation system.]{The figure shows the general structure of a telemanipulation system with five elements; human operator, a input device, communication channel (network), avatar device and remote environment. A human operator using the input device controls/commands the avatar device to interact with/adjust the remote environment. This operation usually occurs through a network (e.g. satellites, relay stations, etc) and is usually slowed down due to bandwidth, network buffering, processing on relay stations or other reasons \cite{iet:/content/books/ce/pbce037e}}.
%     \label{fig:basicstructure}
% \end{figure}
\section{RL Approach to MMT}
\label{section:approach}
\begin{figure*}%%%

\centering
\includegraphics[width=\textwidth]{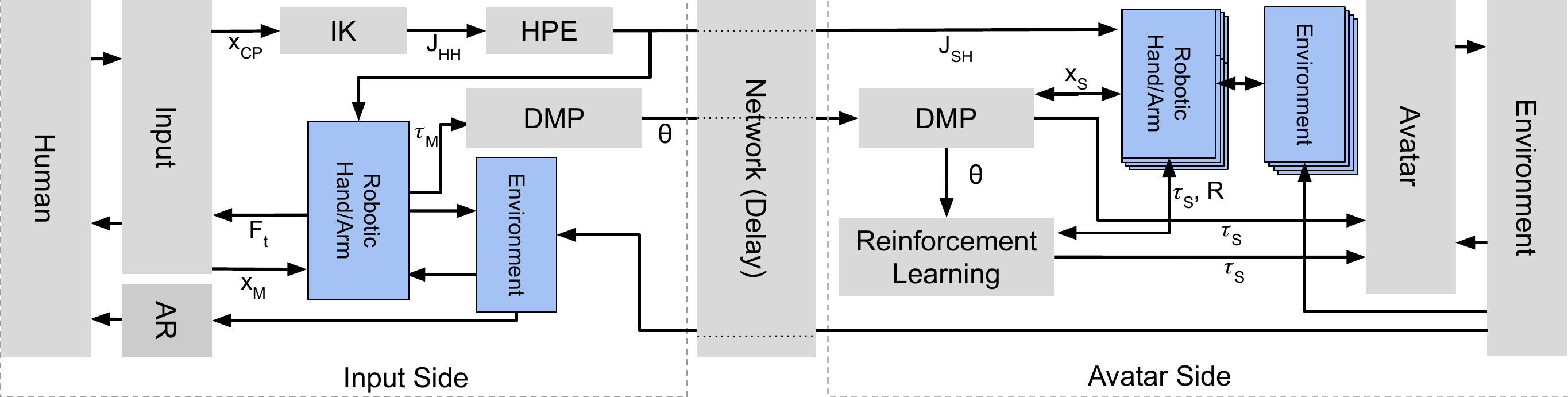}

    \caption[The proposed architecture.]{The overall architecture of the RL enhanced MMT. The network (delay) splits the architecture into two main parts, the avatar and the input side. The \textit{Human} teleoperator interacts with the \textit{input} device. On the input side \textit{LWR/Hand simulation} using physics simulation, \textit{Environment} provides instant haptic feedback and also information for the Augmented Reality unit \textit{AR}. The Inverse Kinematic \textit{IK} block calculates the human hand posture given the contact points on the input device. Consequently, the avatar Hand Posture Estimator \textit{HPE} approximates the corresponding avatar hand configuration. The \textit{DMP} in the input side encodes the trajectory to basis functions while the \textit{DMP} in the avatar side encodes them back, and evaluates and reconstructs the new trajectory based on the new environmental situation given by \textit{Hand/LWR} and \textit{Environment} simulator. In case of failure, the \textit{Reinforcement Learning} unit improves the trajectory and finally deploys it on the \textit{avatar} robot in the remote \textit{Environment}. }
    \label{fig:ProposedArchitecture}
    \vspace{-6mm}
\end{figure*}

%In this section the architecture of the teleoperation system is presented. 
The architecture of the proposed approach is designed in a modular way to facilitate the analysis and testing process. Fig. \ref{fig:ProposedArchitecture} illustrates the overall architecture of our MMT system. The avatar and the input side of the architecture are connected with a communication network that causes a delay in transmission.% Each module (blocks in the figure) is assigned to execute a separate task that is defined as below:

\subsection{Dynamic Movement Primitives (DMP)}
\noindent
Transferring knowledge between structurally different systems requires transformation and usually results in information loss. Similarly, in teleoperation, in most cases, avatar and input have fundamental differences that aggravate a lossless performance. Additionally, information can be lost or altered during the communication process. To evaluate the knowledge transfer and trajectory reconstruction, different properties are taken into account:
 
\begin{enumerate}
    \item Quality of adaptation to new conditions
    % The flexibility of reconstruction and degree of adaptation to new conditions:} due to the time difference between demonstration (input side) and execution (avatar side), the environment may change, the differences might be caused by the avatar robot or an external factor (e.g. slipping, drifting). 
    %Therefore, this dissimilarity should be taken into consideration for new trajectory generation, the potential differences that are taken into account are:
    % \begin{itemize}
    %     \item The starting point of the trajectory
    %         \item The target point of the trajectory
    %             \begin{itemize}
    %     \item Reachability of the target
    %     \item Uncertainty in target position
    %         \end{itemize}
    %     \item Obstacles
    % \end{itemize}
    \item Dimensionality of the encoded space
    % The minimum number of parameters that are required to encode a trajectory. 
    % A low number of parameters leads to,
    %     \begin{itemize}
    %     \item less processing overhead
    %     \item less network traffic
    %     \item faster optimization
    % \end{itemize}
    % To ensure the transformation loss does not affect the system eventually, an optimal trade-off needs to be found as a very high number of parameters leads to complexity and a very low number of parameters leads to inconsistency and excessive loss. 
    \item Safe exploration and learning 
    % The reconstructed trajectory must be as safe as the demonstrated trajectory by the supervising teleoperator. 
    % Since learning mainly relies on exploration in parameter or action-space, intuitive representation of the trajectory helps to keep the exploration safe.   
\end{enumerate}

DMPs can learn to reconstruct new trajectories with different temporal and spatial features using demonstration. Given the position $[x_0, x_1, ..., x_N]$, velocities $[\dot{x_0}, \dot{x_1}, ..., \dot{x_N}]$ and acceleration $[\ddot{x_0}, \ddot{x_1}, ..., \ddot{x_N}]$, DMP parameters $\theta$ can be generated using one-shot learning. 
Each point on the trajectory has six elements, three for position and three for orientation. Since each DMP encodes one dimension, six DMPs are used to encode each demonstrated trajectory. 
% \begin{equation}
%     [\quad x_i \quad y_i \quad z_i \quad \alpha_i \quad \beta_i \quad \gamma_i \quad]
% \end{equation}
The roll, pitch, and yaw of the end-effector are then converted to a rotation matrix by a set of transformations where the singularities are avoided.
The DLR Five-Finger Hand (FFH), which is used as the avatar end-effector, is not considered in the DMP transformation due to the high number of DOF (21 DOF), which is outside of the scope of this paper.

\noindent
As Fig. \ref{fig:ProposedArchitecture} shows, the avatar robot receives the final trajectory from two sources, DMP, and RL units. The DMP in the input side encodes the trajectory to a set of parameters and sends it to the avatar side where another DMP block decodes and rebuilds the trajectory given the updated values from the avatar robot and the environment.

\noindent
Before executing the trajectory on the real robot, a simulation is used to evaluate the trajectory. This paper mainly focuses on grasping different geometrical objects (e.g. centric, parallel, arbitrary); the evaluation conditions are designed correspondingly. If the reconstructed trajectory leads to a successful grasp in simulation, thereafter, the real avatar robot deploys the trajectory. Otherwise, the DMP unit activates the RL to adapt and improve the trajectory until a successful grasp in simulation is achieved.
% \begin{figure}[h!]
%     \centering
%     \includegraphics[width=\linewidth]{images/differentdmp.pdf}
%     \caption[The figure shows the trajectories generated using a different number of kernels.]{The figure shows the trajectories generated using 10, 20, 50, 100 kernels. Increasing the number of kernels improves the accuracy of the trajectory by adding more features throughout the trajectory. on the other hand, the unnecessary high number of kernels increases the learning time and complexity of the trajectory representation. }
%     \label{fig:DifferentnumberofDMPskernels}
% \end{figure}

\subsection{Reinforcement Learning}
Although DMPs can adapt an old trajectory to new conditions, they may fail due to reasons such as the approaching angle for grasping with an asymmetric end-effector structure (e.g. anthropomorphic hand) while for a manipulator with a symmetric end-effector the approach angle does not make any difference. Furthermore, uncertainty in the object position is a common problem and causes collisions and inconsistency in grasping and eventually, failure of the task execution. Hence, a new demonstration by the teleoperator might be costly and also result in the same problem; therefore, our architecture detects such a failure before execution in the real environment and adapts the same trajectory to the new conditions using different reinforcement learning methods.
%There are several ways to improve a trajectory, such as policy search, path integrals, and deep reinforcement learning.
This work presents the comparison between fast and robust algorithms such as Policy Improvement using Path Integrals ($\textrm{PI}^2$), Policy Learning by Weighting Exploration with the Returns (PoWER) and Episodic Natural Actor-Critic (eNAC).
\begin{figure}[b]
% \vspace{-3mm}
    \centering
    \includegraphics[width=\linewidth]{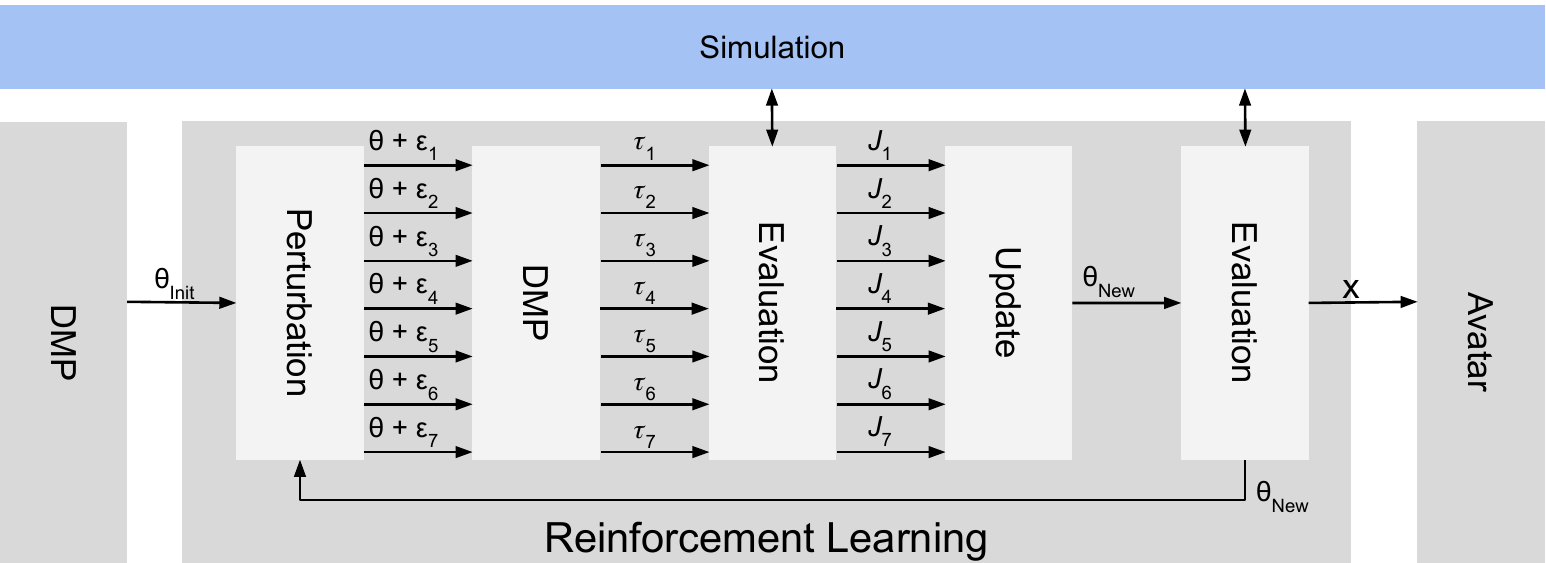}
    \caption[The general structure of policy parameter perturbation methods]{The figure shows the general structure of policy parameter perturbation methods inspired from \cite{freekpi2}.}
    \label{fig:Reinforcementlearning}
\end{figure}

\noindent
$\textrm{PI}^2$ and PoWER are policy perturbation methods that explore the parameter space $\pi_{\theta+\epsilon}$ to generate trajectories that are similar to the demonstration but have slightly different features. These exploratory trajectories are essential for RL to visit new parts of the task-space to find a local optimal solution. Fig. \ref{fig:Reinforcementlearning} shows a detailed view of the RL block in Fig. \ref{fig:ProposedArchitecture}, which outlines the general structure of policy perturbation methods. Both algorithms have a similar exploration behavior, but the differences are in the calculation of parameter update $d\theta$. The parameter update in $\textrm{PI}^2$ is calculated in a way that provides more freedom for designing the cost function. Whereas the cost function in the PoWER algorithm has to be an improper probability distribution, therefore, the returns should always be positive. 

\noindent
The eNAC algorithm \cite{enac} is an actor-critic policy gradient which uses the natural gradient to find the steepest path to the optimal solution. It is called episodic since the critic is evaluated at the end of the trajectory execution, but the algorithm is step-based since the actions are perturbed each frame. The algorithm has several benefits like intuitive exploration rate and using natural gradient but 
perturbing actions has disadvantages \cite{freek:hal-00922132} such as,
\begin{itemize}
    \item Introducing independent random small movements to a smooth trajectory increases the jerkiness,
    \item The system might behave as a low-pass filter and filter-out the high-frequency perturbations by averaging.
    \item Instant changes in robot motion without considering the inertia might damage the robot,
    \item The variance of the parameter vector might increase dangerously.
\end{itemize}
The details regarding the formulas and algorithms can be found in \cite{enac, MainPI2,PowerAlgorithm}.

\subsection{Learning under Uncertainty}
Due to the uncertainty in object detection, the RL agent must be able to learn the new trajectory goal while optimizing the trajectory shape. 
The authors in \cite{freekpi2} introduce an approach to extend the capability of $\textrm{PI}^2$ to adapt the target of the trajectory. The approach uses the cost of the whole trajectory since the goal does not change throughout the trajectory execution. The haptic feedback as secondary sensory feedback is used to compensate for the vision inaccuracy. 

\noindent
\subsection*{Formulation}
\subsubsection{Exploration Rate $\Sigma$}
As $\textrm{PI}^2$ and PoWER approaches suggest, the perturbation step introduces exploration into the parameter space. The exploration may cause serious disruptions in the agent's behavior and unpredictable movements, which can adversely affect the avatar robot and the remote environment simultaneously. Due to the counter-intuitive definition of the parameters $\theta$, which define the final shape of the trajectory by modifying the DMPs, an intuitive understanding of the applied noise (exploration) $\Sigma$ is required before performing actions on the avatar. A high exploration rate in parameter perturbation methods may highly diverge the trajectory and cause physical damage to the robot.

\noindent
As mentioned, the eNAC algorithm uses the action-space to explore the new policies; therefore, the exploration rate $\Sigma$ is an intuitive measure of distance. A low exploration rate leads to a smoother trajectory, while a high exploration rate may cause damage to the robot by big instant changes. In parameter and action-space perturbation methods, a trade-off can be found that speeds up learning while keeping the avatar device and remote environment safe.  

\noindent
Furthermore, initially, the RL agent requires higher exploration to increase the chance to find a solution within the time limit. For that reason, when a solution has been found, the rate of the exploration must decrease to stabilize the convergence and finalize the trajectory shape. Therefore, the exploration rate $\Sigma$ decreases at each time step using the following equation:
\begin{equation}
\gamma = max(\cfrac{\textrm{Update}_{\textrm{max}} - i}{\textrm{Update}_{\textrm{max}}}, 0.1).
\end{equation}
\begin{equation}
\Sigma_i = \gamma \Sigma_{\textrm{init}}.
\end{equation}
Where $i$ is the current number of updates and $\textrm{Update}_{\textrm{max}}$ is the maximum number of updates. The decay rate $\gamma$ is responsible for linearly decreasing the exploration rate.
% The decay rate has a minimum value of 0.1 due to need for exploration throughout the learning process. 
On the other hand, the exploration rate of the goal defines the perturbation of the target position.
% Therefore high goal exploration leads to failure due to diverging quickly before having a chance to touch the object. On the other hand, a low exploration rate results in late convergence, therefore the trade-off should be defined properly depending on the model uncertainty.
The learning exploration rate for each approach was defined experimentally in the simulation (see table \ref{tab:exploration_rate}).
\begin{table}[b]
    \caption{The table shows the exploration rate defined for each approach}
    \label{tab:exploration_rate}
    \centering
        \begin{tabular}{|l|l|l|l|l|}
            \hline
               & $\textrm{PI}^2$ & PoWER & eNAC & Goal\\ \hline
            $\Sigma$ & 300 & 300 & 0.01 & 0.04\\ \hline
        \end{tabular}
\end{table}

\subsubsection{Learning Rate $\alpha$}
In the critic step of the eNAC algorithm, a learning rate is deployed to control the final influence of parameter updates on the final trajectory. A small learning rate leads to late convergence, while a high learning rate results in sudden changes in the trajectory and failure due to high divergence from the initial trajectory.  
\subsubsection{Cost Function}
The cost function in policy search algorithms defines whether the agent should get a reward. It also defines the scale of the reward, depending on the quality of the action. A cost function maps several aspects of action in one single value with different proportions depending on the purpose of the learning. For example, the cost function can be a combination of:
\begin{itemize}
    \item Acceleration or velocity of the end-effector
    \item Distance from the target
    \item The scale of the exploration noise
    \item The number of frames that fingertips are in proximity of an object
    \item The number of fingers involved in the grasp
    \item The object displacement during grasp
\end{itemize}   
\begin{figure*}
\centering
\includegraphics[width=\textwidth]{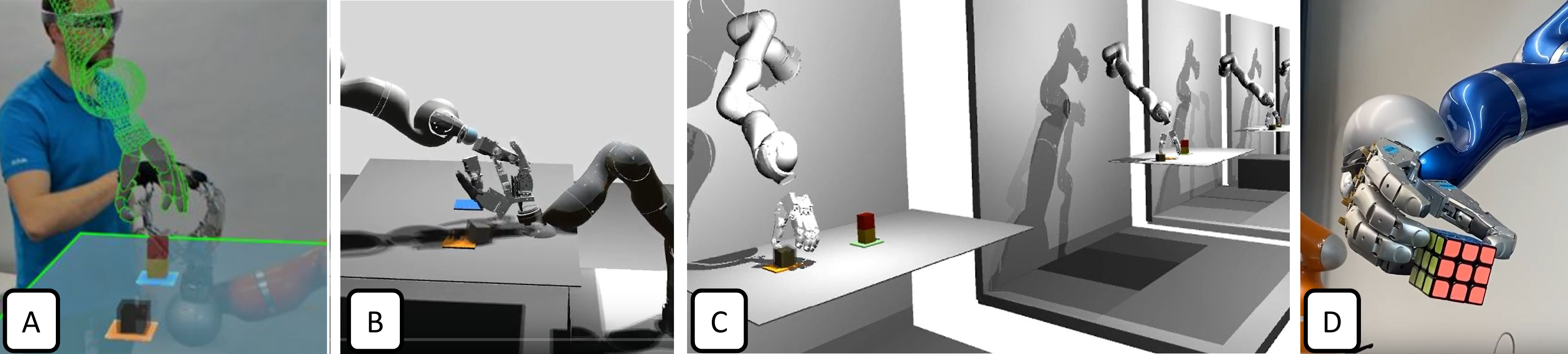}
    \caption{The learning procedure for grasping a cube. In A, the teleoperator uses the Exodex Adam haptic interface to teach the avatar robot to grasp a cube. In B, a simulation of avatar robot and input setup processes all the physical properties of the environment like forces e.g. gravity and lets the human interact with objects in real-time. In C, multiple avatar units try different approaches to find a solution in case the environment has changed during data transmission. In D, finally, the evaluated trajectory in the simulation is used to grasp the object in the real avatar setup.}
    \vspace{-6mm}
    \label{fig:SystemInReal}
\end{figure*}
\noindent
Different combinations of these parameters have a different interpretation, and they result in different trajectories. For example, integrating the acceleration into the cost function results in a less jerky trajectory \cite{freekpi2}.
%Another example, by adding the object displacement during the grasp makes the end effector not touching during its approach. 
%Consequently, considering any mentioned property results in a new trajectory where that property is minimized. 
Moreover, The simulation involves characteristics that are unrealistic, like continuous access to the precise object position. In a real environment, the object can be obscured and hard to detect and locate; therefore, using cost items like the object position is not plausible. As a final solution, two properties are experimentally selected; acceleration and noise scale for the control and trajectory cost. And for the final cost, the number of fingers that have touched the object is included. 
The cost function is defined as below \cite{freekpi2},
\begin{equation}
    J(\tau_i) = \Phi_{t_{N}} + \int_{t_i}^{t_N} (10^{-11}(\ddot{x_t}^2 + \frac{1}{2} \theta_t^T R\theta_t)) dt.
\end{equation}
where $\Phi$ is the final cost indicating the quality of the grasp. The quality of the grasp is determined by the number of fingers involved in the grasp $N_{Fingers}$,
\begin{equation}
    \Phi(\tau_i) = 1 - \frac{ N_{Fingers} }{ 5 }.
\end{equation}

\noindent
The final cost $\Phi$ is zero when all fingers are involved in the grasp, but depending on the object size, the maximum number of fingers might differ.

\subsubsection{Roll-out and Update number}
The number of roll-outs determines the number of perturbed trajectories which must be generated and evaluated for calculating the next update. The number of updates defines the maximum number of times that the policy parameters get updated.
The higher number of roll-outs leads to robust learning but slower convergence \cite{freekpi2}. Maximum 100 updates with seven roll-outs have been used in the experiments due to hardware limitations. To increase the learning stability, the best two trajectories are kept in the memory to be used in the next update.

%\section{Implementation and Test-bed}
\section{Teleoperation System Implementation}
\label{section:implementation}

To provide a framework to evaluate the system, different modules are designed and implemented. 
%The main focus of this chapter is the link between the hardware and software, controllers, and description of remained modules. The hardware architecture including manipulators, robotic hands, visualization devices, and the middle-ware are explained with a brief description of the experiments. 
%As Fig. \ref{fig:SystemInReal}.A shows, the input device, DLR Exodex Adam, a novel haptic interface designed by the MODEX lab at German Aerospace Center \cite{p:Hapi,p:Exodex Adam,p:Exodex-Prime}.
Fig. \ref{fig:first_image} and Fig. \ref{fig:SystemInReal}.A show the input device, DLR Exodex Adam, a novel haptic interface designed by the MODEX lab at German Aerospace Center \cite{p:Hapi,p:Exodex-Adam,p:Exodex-Prime}.
%\color[rgb]{0,0,1}\textbf{Add a figure with the system here. You can use any picture with a person operating the system, if possible, with the Hololens on the user. Also show what the user sees if possible. Make sure the caption describes all of this.}.
This is paired with a Microsoft Hololens as the AR to provide a visualization of the virtual environment as an overlay on the operator’s vision. The augmented 3D models of the table, objects, and robots are processed in Unity game engine\protect\footnote{Unity game engine available at \protect\url{https://unity.com}} and visually anchored using Vofuria engine\protect\footnote{Vofuria engine available at \protect\url{https://engine.vuforia.com/engine}}. The avatar setup uses a robot arm and a customized version of the DLR Five-Finger Hand (FFH)\cite{taskman}. The differences between The FFH and DLR hand is in the position of the thumb, which is opposing other fingers to facilitate grasping and in-hand manipulation \cite{dlr93720}. The avatar robot arm is a custom-configured LWR 4+, with physical offset modifications on joints 1, 3, and 6, to facilitate the anthropomorphic configuration. The Links and Nodes (LN) library is used as middleware to manage the processes and the communication between all the components/agents.

\noindent
High-level task demonstration, such as grasping in a teleoperation scenario, requires a highly compliant interface that provides high flexibility and weightless movement~\cite{aaron_adaptive_grasping}. In addition to compliance, proper haptic feedback helps the teleoperator to understand the physical properties of the remote environment by interacting with the objects. To do so, a gravity compensation and a regular torque controller were deployed on the input device to provide compliance and flexibility on the arm and hand (fingers).
% Consequently, the in-hand (e.g. gestures) and human arm movements were followed by the robot fingers and the manipulator, correspondingly by the avatar robot.
On the avatar-side, the robot requires a position to follow the input; this position, called anchor, can be any point on the input device. For instance, the human hand palm position or the input's end-effector position would are proper options. The anchor is a virtual link that the avatar robot uses to follow the input movements.
A pipeline of Inverse Kinematic (IK) \cite{aaron_adaptive_grasping} and joint-to-joint mapping has been used to estimate the human hand posture using the avatar hand.
The IK is used since there is no direct way to read human joint values. It utilizes the contact points calculated using forward kinematic on the input device. Then, the joint-to-joint mapping uses the human hand joint configuration to compute the avatar robot joint values using optimization.
 Since MMT extensively uses a simulation environment, therefore using a simulation with a proper physics engine is necessary. The chosen physics engine must have the ability to:
 \begin{itemize}
    \item Calculate penetrations on complex objects
    \item Simulate rigid body dynamic behavior
\end{itemize}
Unity is a cross-platform game engine with a powerful built-in physics engine. Although Unity is a powerful tool to simulate complicated objects with dynamic behavior but due to limited processing power, the capabilities are restricted. For example, Unity does not support non-convex rigid body collision detection due to computation overhead and low demand in the market. 
%%Final Cut%% An object is convex when there is no line inside the object that intersects with its surface mesh. Consequently, Unity has a default approach to deal with non-convex objects which is to approximate a new convex-mesh so-called the collider. 

\noindent
Although Unity supports simulation for different tangential forces for instance friction, the positional noise coming from the input robot makes the surface friction impractical. Surface friction is necessary for grasping; this tangential force is the main reason that the object stays within the hand during the movements. In the simulation, due to the noise from the real input device, the surface friction fails because of constant attachment and detachment of the fingertips and the object surface.
This fluctuation causes the object to slide out of the hand right after picking up off the ground. To solve this problem using Unity a new approach is proposed so-called the diaphragm. The diaphragm guarantees a smooth attachment without any contact issue.
The diaphragm is a novel idea which uses an arbitrary virtual object surrounding the target object to adapt contact forces between the hand and the objects. It has the same shape as the object but 20 percent bigger in size with a transparent appearance. Once one of the robot fingers enter the diaphragm, the physics engine takes over the low-level control. Although the diaphragm facilitates the grasping, still due to high noise in object position, it may slip out of the hand, to address this issue an external controller is deployed on the object position. This external controller makes the grasp more persistent by using a kinematic control and eliminating all the forces, including gravity. The depth and normal vector of the penetration calculated in simulation are sent to the input robot, where the god object approach \cite{godobjects1} is used to provide haptic feedback for the operator.

\section{Results}
\label{section:results}
\begin{figure}[t]
    \centering
    \includegraphics[width=\linewidth]{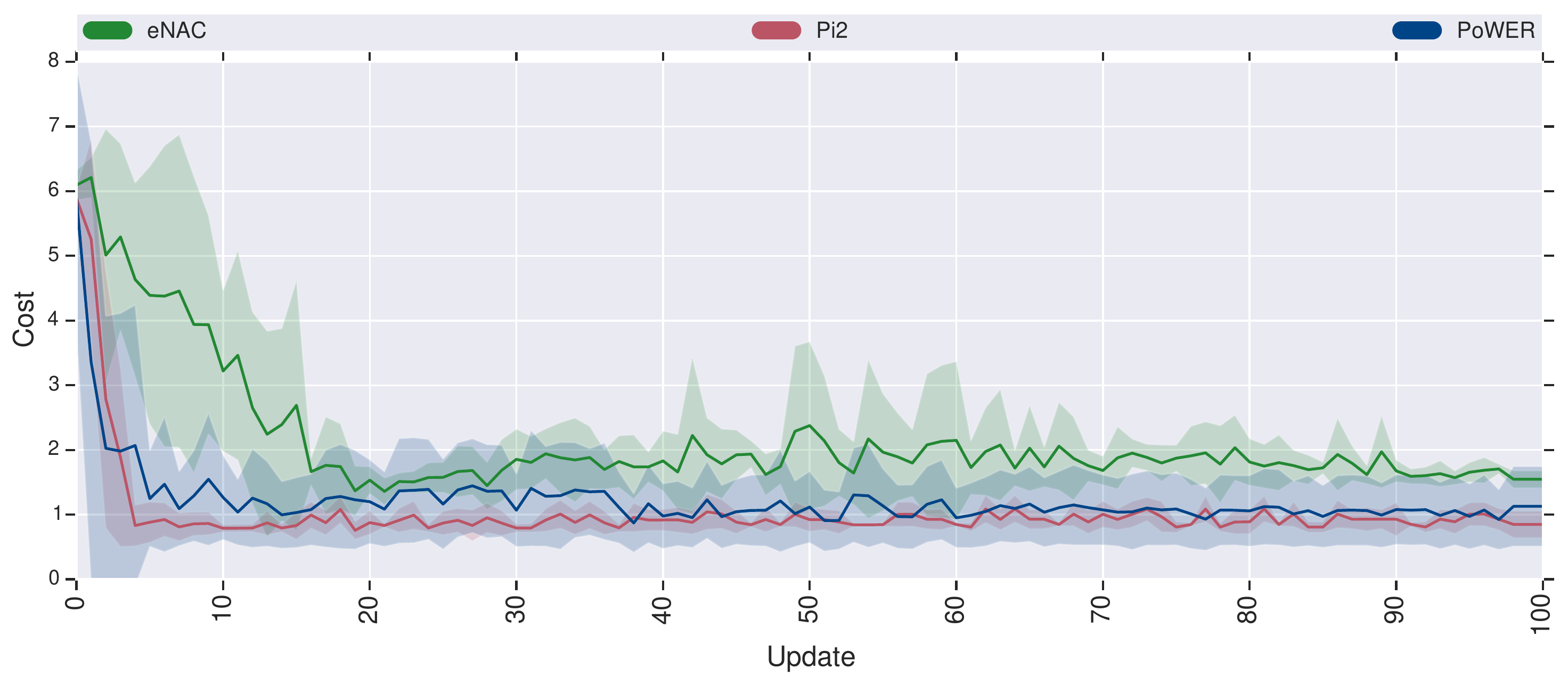}
        %   \vspace{-8mm}
    \caption[The learning cost over 100 updates.]{The learning cost over 100 updates. Each update contains seven roll-outs performed in the simulation. The targets related to this plot were placed about 50 centimeters away from the demonstration. The figure shows $\textrm{PI}^2$ and PoWER has a very steep decrease in the cost of the trajectory while eNAC has a slower convergence and less stability with high fluctuations.}
    \label{fig:LossRateBox}
    \vspace{-3mm}
\end{figure}

\begin{figure}[b]
  \centering
  \includegraphics[width=\linewidth]{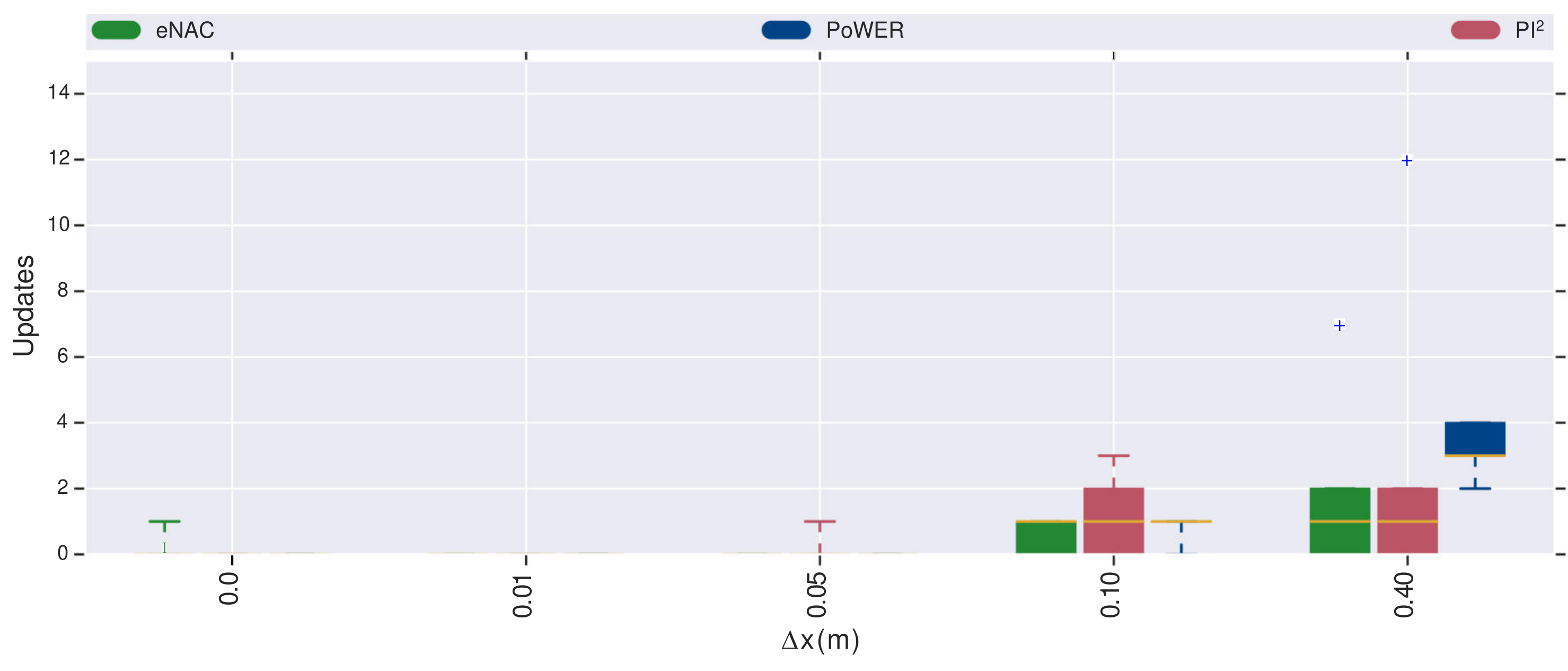}
%   \vspace{-8mm}
  \caption[The update number required for different approaches to find a solution.]{The update number required for different approaches to find a solution. The horizontal axis indicates the distance from the demonstrated target, and the vertical axis shows the number of updates. The positional differences are applied to the X direction.}
      \label{fig:NumberUpdatesBoxX}
      \vspace{-3mm}
\end{figure}
\begin{figure}[h]
   \centering
    \includegraphics[width=\linewidth]{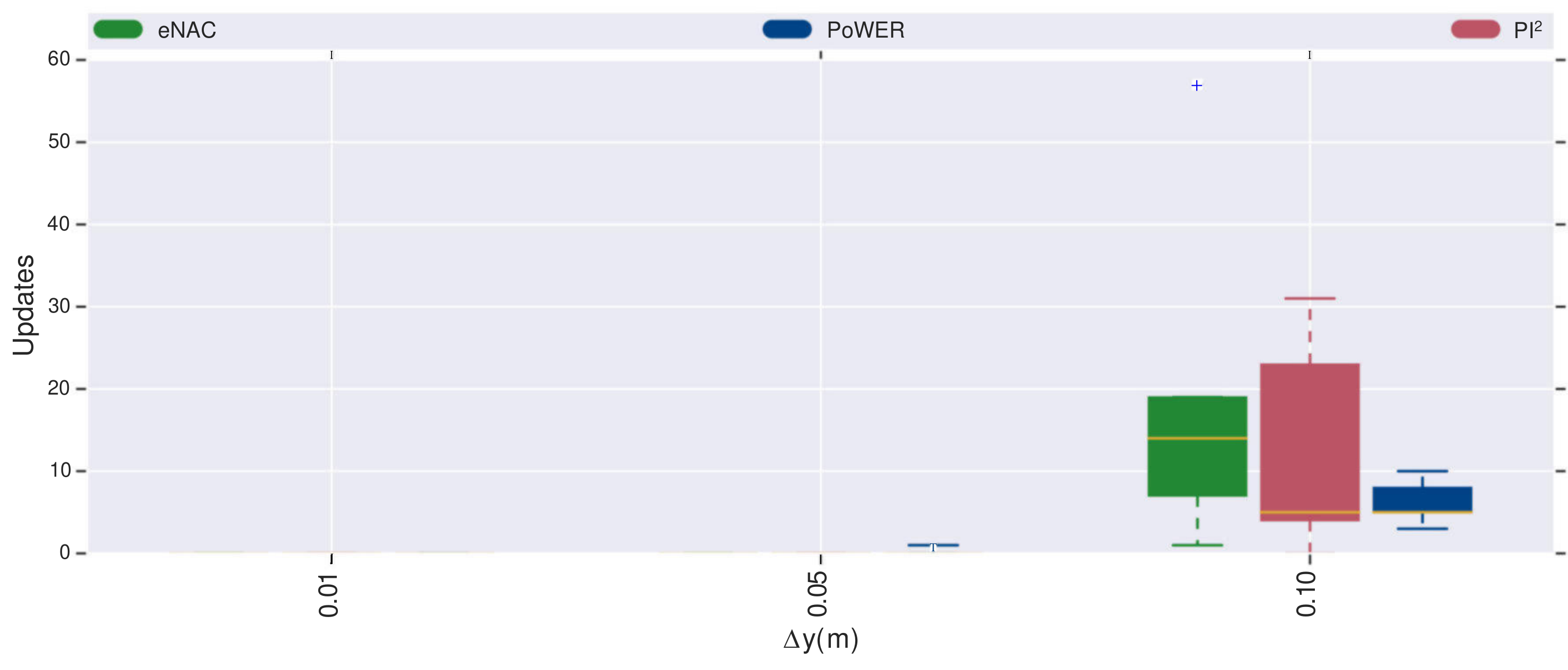}
    %   \vspace{-8mm}
    \caption[The update number required for different approaches to find a solution.]{The update number required for different approaches to find a solution. The horizontal axis indicates the distance from the demonstrated target, and the vertical axis shows the number of updates. The positional differences are applied to the Y direction.}
      \label{fig:NumberUpdatesBoxY}
    \vspace{-3mm}
\end{figure}

\begin{figure}[b]
    \centering
        \vspace{-3mm}
     \resizebox {\linewidth} {!} {
    \input{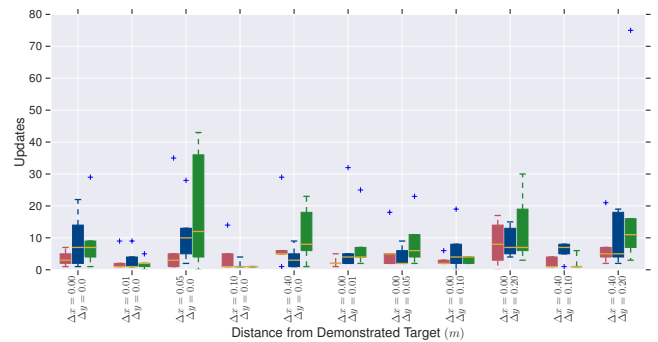}
    }
                %   \vspace{-8mm}
    \caption[Comparison between different approaches in grasping the cylinder.]{Comparison between  different approaches in grasping the cylinder. The result shows a small deviation from the original position of the object during demonstration results in a DMP failure. $\textrm{PI}^2$ shows the best performance, while eNAC requires the highest number of updates.}

    \label{fig:Certain_Cylinder}
\end{figure}

%%Final_Cut_Temp%
% \begin{figure}
%     \centering
%      \resizebox {\linewidth} {!} {
%     \input{images/Paths_V1.pgf}
%     }

%     \caption[The trajectories generated by different approaches.]{The trajectories generated by different approaches. In $T_x$, the eNAC trajectories show a high divergence from the demonstrated trajectory. The trajectories generated using $\textrm{PI}^2$ have the lowest deviation from the demonstration. }
%         \label{fig:Paths_V1}
% \end{figure}

%%%%%%%%%%%%%%%%%%%%%%%%%%%%% Certain BOX Cost %%%%%%%%%%%%%%%%%%%%%%%%%%%%%%%%%%%%%%%%%%%%%%%%%%%%
The first set of results compare different algorithms in a scenario where the avatar robot grasps a box. Fig. \ref{fig:LossRateBox} compares the cost of trajectories generated by different methods in a condition where the object is located almost 40 centimeters away from the demonstration in the XY plane (table top). 

\noindent
The initial cost relates to the trajectory generated by DMPs, and later, each algorithm attempts to reduce costs within a limited time. The figure shows that all methods successfully decrease the cost and stabilize the learning with a low cost where the grasping happens. The $\textrm{PI}^2$ and PoWER rapidly decrease the cost, but $\textrm{PI}^2$ shows higher stability by less fluctuation while reaching a lower value in the steady-state. The eNAC algorithm has a gradual decrease and unusual fluctuations in the steady-state. Moreover, eNAC's steady-state shows a higher final cost than the other approaches.

\noindent
Fig. \ref{fig:NumberUpdatesBoxX} and Fig. \ref{fig:NumberUpdatesBoxY} compare the number of required updates to achieve a successful grasp when the object is moved in the XY Plane. Moreover, zero updates indicate the DMPs have compensated for the changes, and the RL was not used. The box charts indicate the results gathered from five different learning sessions for each algorithm. The figures show the changes in Y are harder to compensate for the DMPs due to workspace limitation and robot hand kinematics.

\noindent
As shown, the RL is required when the changes are greater than ten centimeters on the X-axis, where the median of the required updates is one. When the change reaches 40 centimeters, the PoWER algorithm needs more updates, and the outliers show eNAC and $\textrm{PI}^2$ encounter occasional difficulties to find a solution. When applying changes in the Y-axis, the PoWER algorithm has a better performance by having a smaller inner fence and lower median. The $\textrm{PI}^2$ has the same average but has a wider boundary. Additionally, eNAC has the same issue plus a higher median and also an outlier, which shows the algorithm needed more than 50 updates to find a proper solution. 

%Final_Cut_Temo% Fig. \ref{fig:Paths_V1} shows the final trajectories generated using different algorithms. Trajectories have been calculated with no perturbation after the last update. The figure shows eNAC drastically changes the trajectory in $T_X$ and $T_Z$ while PoWER aimed at finding a solution by applying changes in $T_Z$ and $R_X$. The figure shows that $\textrm{PI}^2$ has the least modification to the demonstrated trajectory in all axes. The same results are achieved by~\cite{pmlr-v9-theodorou10a}.

Fig. \ref{fig:Certain_Cylinder} compares the number of required updates for grasping a cylinder with different approaches. The result shows a small deviation from the original position of the object during demonstration results in a DMP failure, which made this task the hardest among the others. The results also show $\textrm{PI}^2$ has the best performance while eNAC requires the highest number of updates.

\begin{figure}[]
    \centering
     \resizebox {\linewidth} {!} {
    \input{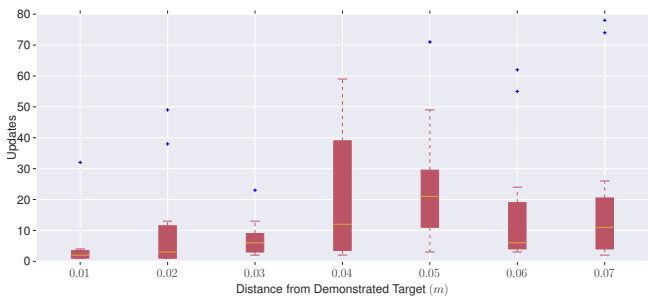}
    }
            %   \vspace{-8mm}
    \caption[The number of necessary updates for $\textrm{PI}^2$ shape and goal learning under uncertainty.]{The number of necessary updates for $\textrm{PI}^2$ shape and goal learning under uncertainty. The vertical axis determines the number of updates, and the horizontal axis determines the magnitude of uncertainty. The magnitude increases to seven centimeters from the initial object position. The result shows an increase in the number of updates due to increasing the uncertainty, but an anomaly has happened after five trials, and the number of total updates decreases, which can be attributed to the exploration rate.  }
    \label{fig:Uncertainty_Num}
\vspace{-6mm}
\end{figure}

\noindent
Fig. \ref{fig:Uncertainty_Num} compares the number of updates required to achieve a successful grasp under uncertainty. As mentioned, although DMPs are powerful tools, but they fail with small uncertainty in the object position. By increasing the uncertainty, the number of required updates increases as expected. The medians in the figure show that the number of required updates decreases when the uncertainty reaches six centimeters. The drop might be due to the goal exploration rate. So the end-effector does not hit the object, since initially, it is not in the exploration range. Therefore, smaller exploration rates were evaluated, but the results were not satisfying, and the RL agent mostly failed to reach the object in 100 trials. The failure is due to the low probability of finding the object since the haptic feedback is used as a reward, so as long as the object is not touched by at least one of the fingers, the agent does not receive any reward. For example, When the uncertainty is six centimeters, but the exploration rate is two centimeters the agent needs at least three trials toward the object with maximum step size to gain reward from touching the object, which is quite unlikely to happen. 
% \begin{figure}
%     \centering
%      \resizebox {\linewidth} {!} {
%     \input{images/ena2.pgf}
%     }
%         \label{fig:differentSteps}
%     \caption[The figure compares different step sizes in the eNAC algorithm.]{The figure compares different step sizes in the eNAC algorithm. The figure shows step size 10 has the best performance and requires fewer updates. The vertical axis determines the number of updates and the horizontal axis determines the magnitude of change in the object position relative to the demonstration. }
% \end{figure}
%%%%%%%%%%%%%%%%%%%%%%%%%% Uncertainty Trajectory %%%%%%%%%%%%%%%%%%%%%%%%%%%%%%%%%%%%%%%%%%%%%%%%%%%%%%%

\begin{figure}[t!]
    \centering
    \includegraphics[width=\linewidth]{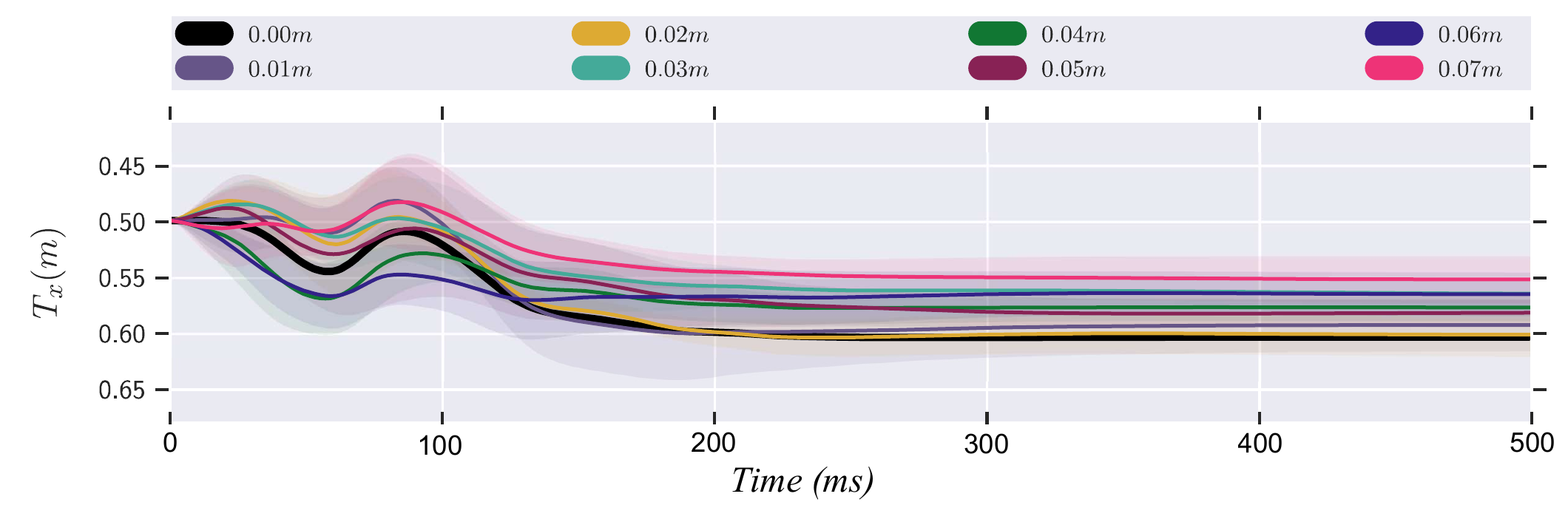}
    \caption[The trajectories generated under different uncertainty in object position.]{The trajectories generated under different uncertainty in object position. The vertical axis shows the translation of the end-effector pose in X axis, and the horizontal axis determines the time. }
    \label{fig:Uncertainty_Path}
    \vspace{-5mm}
\end{figure}

\noindent
Fig. \ref{fig:Uncertainty_Path} shows the trajectories generated under different uncertainties in object position. As the trajectories illustrate, the agent has successfully found several solutions for each trial with different uncertainties. The final stage of each trajectory shows where the end effector has grasped the object. The final values of the trajectories, in comparison with the final values of the demonstration, show the approach has successfully found a solution.

%%% Final cut% For example, the generated trajectory has the largest deviation when the uncertainty is seven centimeters. Since the robot hand is relatively big, the trajectories with uncertainties less than 3 centimeters are not corresponding to their plots. For example, the trajectory generated for 2 centimeters uncertainty has less deviation at the end than the trajectory generated for 1 centimeters uncertainty.

\section{Discussion}
\label{section:discussion}
DMPs can help cope with challenging problems such as the ball-in-cup game, obstacle avoidance, and grasping \cite{mastersthesis1, 4755937, MATSUBARA2011493}. The results from other research \cite{freekpi2} have also shown the method to perform well in a real uncertain environment. In this work, the complexity and challenge of the problems stem from the integration of DMPs into a teleoperation framework. Furthermore, since uncertainty in many cases results from the use of physical hardware/avatar devices, it is essential to evaluate the approach under real conditions. %The results from other research \cite{freekpi2} already show the method performs well in a real uncertain environment. But, this paper evaluates the result in a new hardware setup, simulation set, and different conditions. The main reason that the learning on the real robot was not realized was the time limit due to the relatively high number of updates and roll-outs required for a successful grasp. 
The average delay time of the operation process depends on the demonstration time, network delay, environmental changes (real-time DMP) and the uncertainty of the task (non-real-time RL). The user, however, does not experience this delay as haptic feedback is generated in real-time based on the local model of the avatar environment.
This paper aims to evaluate the system in a simulation environment and assess the ability to use sim-to-real where learning happens in simulation and the final results are executed in the real system to speed up the process. The smooth transfer from simulation to real-world will depend on several factors, e.g., correct camera calibration. Our approach takes advantage of combining MMT and learning from demonstration.
Furthermore, the algorithms and approaches that have been used in this study were all policy search methods that explore the state-space locally. Therefore, it is very likely that the generated solutions are around the local minimum. The deep RL approaches use a global exploration method, which can improve the generality of the final solution \cite{Hadi_ICANN, Matthias_IJCNN}. 

\noindent
However, deep RL uses neural networks as policy and value functions, and this requires a lot of training data. Therefore, if the avatar robot spends a long time gathering data, the teleoperation may fail, or the inconsistency between demonstration and the environment may increase. Augmented Reality (AR) is used with the intention to give the operator visual awareness of the input device while performing the tasks. However, some operators/subjects have expressed that this caused some distraction. With this in mind, a VR interface may be considered to completely occlude the input device from the operator's line of sight. 

\section{Conclusion}
\label{section:conclusion}
%We demonstrated that the MMT architecture integrated with DMPs can cope with 
We present a novel approach to Model-Mediated Teleoperation for grasping and manipulation, and show that integrating reinforcement learning and DMPs in the control can cope with challenging problems in a long-distance teleoperated grasping scenario under long time-delays. We also show $\textrm{PI}^2$ has the best performance to adapt to new conditions under high uncertainties in the model. Our approach is realized and examined on DLR's Exodex Adam to operate in a simulated environment through Microsoft Hololens to help assess the ability to use sim-to-real to speed up the process. 
We believe that apart from looking for learning from demonstration, future research should look for different methods such as learning from imitation and from videos to facilitate the learning process.
Regardless, future research could continue to explore the online Deep Reinforcement Learning (DRL) methods, and investigating the effect of using a virtual reality representation might prove important.
%%%%%%%%%%%%%%%%%%%%%%%%%%%%%%%%%%%%%%%%%%%%%%%%%%%%%%%%%%%%%%%%%%%%%%%%%%%%%%%%
    \bibliographystyle{unsrt}             % Style for presenting the literature
    \addcontentsline{toc}{chapter}{Bibliography}% Add to the TOC
    \bibliography{thesis}
%\end{thebibliography}
\end{document}